\documentclass{article}

\usepackage{microtype}
\usepackage{graphicx}
\usepackage{subfigure}
\usepackage{booktabs} 

\usepackage{hyperref}

\usepackage[accepted]{icml2018}

\icmltitlerunning{Alpha-Beta Variational Inference}

\usepackage{JB_config_article}
\graphicspath{{Images/}}

\begin{document}

\twocolumn[
\icmltitle{Alpha-Beta Divergence For Variational Inference}



\icmlsetsymbol{equal}{*}

 \begin{icmlauthorlist}
  \icmlauthor{Jean-Baptiste Regli}{ucl}
  \icmlauthor{Ricardo Silva}{ucl}
\end{icmlauthorlist}

\icmlaffiliation{ucl}{University College of London, London, United Kingdom}

\icmlcorrespondingauthor{Jean-Baptiste Regli}{ucakgar@ucl.ac.uk}

\icmlkeywords{variational inference, divergence measures, Bayesian neural networks}

\vskip 0.3in
]



\printAffiliationsAndNotice{}    

\begin{abstract}
	This paper introduces a variational approximation framework using direct 
optimization of what is known as the {\it scale invariant Alpha-Beta divergence} 
(sAB divergence). This new objective encompasses most variational objectives 
that use the Kullback-Leibler, the R{\'e}nyi or the gamma divergences. It also 
gives access to objective functions never exploited before in the context of 
variational inference. This is achieved via two easy to interpret control 
parameters, which allow for a smooth interpolation over the divergence space 
while trading-off properties such as mass-covering of a target distribution and 
robustness to outliers in the data. Furthermore, the sAB variational objective 
can be optimized directly by repurposing existing methods for Monte Carlo 
computation of complex variational objectives, leading to estimates of the 
divergence instead of variational lower bounds. We show the advantages of this 
objective on Bayesian models for regression problems.
\end{abstract}

\section{Introduction}
  \label{sec:Introduction}
  
  Modern probabilistic machine learning relies on complex models for which the 
exact computation of the posterior distribution is intractable. This has 
motivated the need for scalable and flexible approximation methods. Research on 
this topic belongs mainly to two families, sampling based methods constructed 
around Markov Chain Monte Carlo (MCMC) approximations~\cite{robert2004monte}, or 
optimization based approximations collectively known under the name of 
\emph{variational inference} (VI)~\cite{jordan1999introduction}. In this paper, 
we focus on the latter, although with the aid of Monte Carlo methods.

\begin{figure}[h]
    \includegraphics[width=\linewidth]{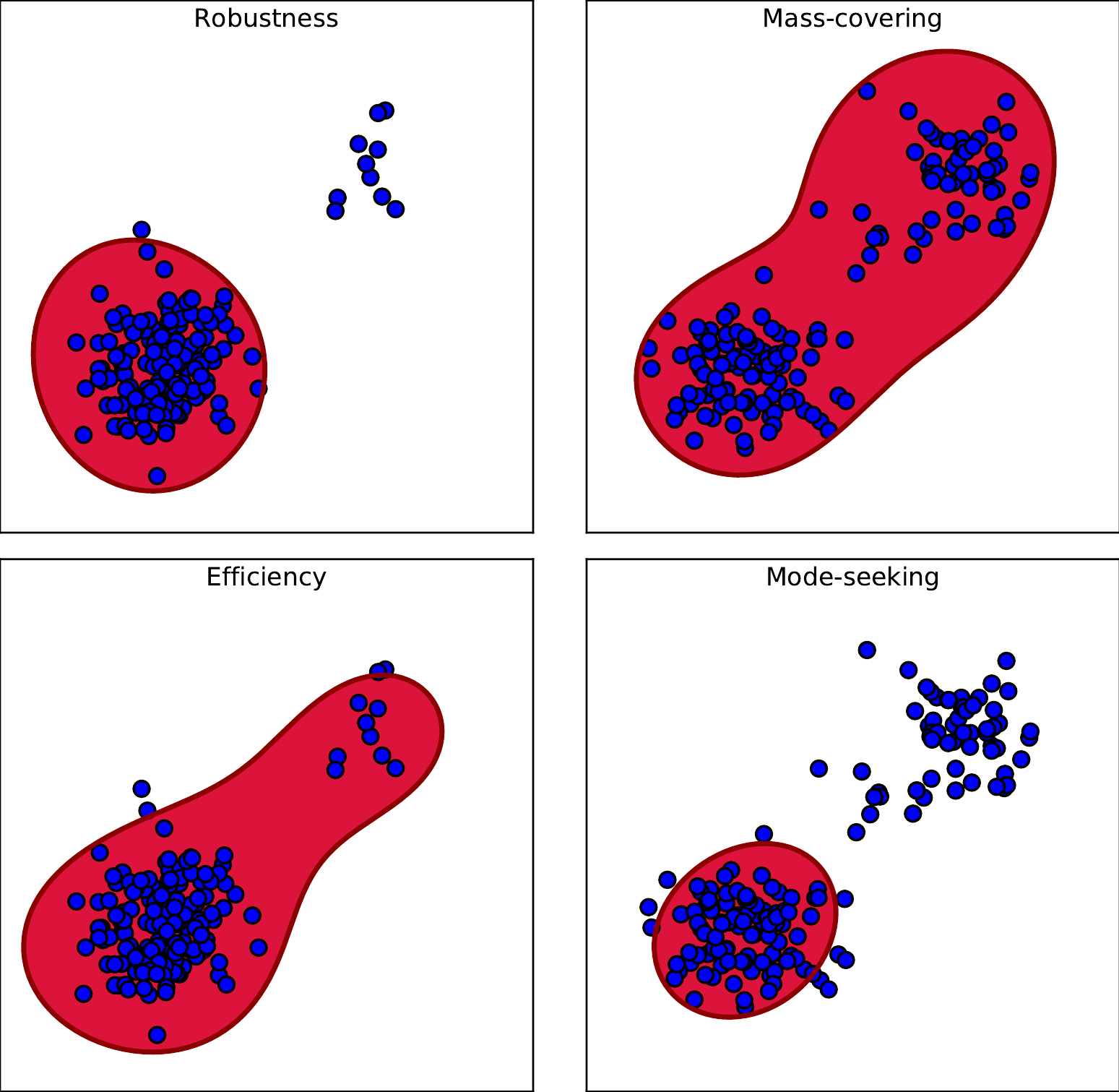}
    \caption{Illustration of the robustness/efficiency properties (left) and 
mass-covering/mode-seeking (right). The red region is a stylized representation 
of a high probability region of a model approximated to fit training data (blue 
points). Mass-covering and mode-seeking are well-established concepts described 
by~\cite{minka2005divergence}.  Efficiency refers to the ability of capturing 
the correct distribution from data, including tail behavior. Robustness is 
defined here as the ability of ignoring points contaminated with noise that are 
judged not to be representative of test-time behavior if their probability is 
too small, according to a problem-dependent notion of outliers.}
    \label{fig:robustness_masscovering}
  \end{figure}

  The quality of the posterior approximation is a core question in variational 
inference. When using the   KL-divergence~\cite{kullback1951information} 
averaging with respect to the approximate distribution, standard VI methods such 
as mean-field underestimate the true variance of the target distribution. In 
this scenario, such behavior is sometimes known as 
\emph{mode-seeking}~\cite{minka2005divergence}. On the other end, by 
(approximately) averaging over the target distribution as in 
Expectation-Propagation, we might assign much mass to low-probability 
regions~\cite{minka2005divergence}. In an effort to smoothly interpolate between 
such behaviors, some recent contributions have exploited parameterized families 
of divergences such as the alpha-divergence~\cite{amari2012differential, 
minka2005divergence, hernandez2016black}, and the 
R{\'e}nyi-divergence~\cite{li2016renyi}. Another fundamental property of an 
approximation is its \emph{robustness to outliers}. To that end, divergences 
such as the beta~\cite{basu1998robust} or the 
gamma-divergences~\cite{fujisawa2008robust} have been developed and widely used 
in fields such as matrix factorization~\cite{fevotte2011algorithms, 
cichocki2010families}. Recently, they have been used to develop a robust pseudo 
variational inference method~\cite{futami2017variational}. A cartoon depicting 
stylized examples of these different types of behavior is shown in 
Figure~\ref{fig:robustness_masscovering},
  
  We propose here a variational objective to simultaneously trade-off effects of 
mass-covering, spread and outlier robustness. This is done by developing a 
variational inference objective using an extended version of the alpha-beta (AB) 
divergence~\cite{cichocki2011generalized}, a family of divergence governed by 
two parameters and covering many of the divergences already used for VI as 
special cases. After reviewing some basic concepts of VI and some useful 
divergences, we extend it to what we will call the scale invariant AB (sAB) 
divergence and explain the influence of each parameter. We then develop a 
framework to perform direct optimization of the divergence measure which can 
leverage most of the modern methods to ensure scalability of VI. Finally, we 
demonstrate the interesting properties of the resulting approximation on 
regression tasks with outliers.

\section{Background}
  \label{sec:Background}
  This section briefly reviews the basis of variational inference. It also 
introduces some divergence measures which have been used before in the context 
of VI, and which will be used as baselines in this paper.
  
  \subsection{Variational Inference}
    \label{subsec:Background/VI}
    We first review the variational inference method for posterior 
approximation, as typically required in Bayesian inference tasks. Unless stated 
otherwise, the notation defined in this section will be used throughout this 
document.
    
    Let us consider a set of $N$ i.i.d samples $\bfX = \{\bfx_{n} \}_{n=1}^{N}$ 
observed from a probabilistic model $p(\bfx|\theta)$ parametrized by a random 
variable $\theta$ that is drawn from a prior $p_{0}(\theta)$. Bayesian inference 
involves computing the posterior distribution of the unknowns given the 
observations:
    \begin{equation}
      \label{eq:Bayes_update}
      p(\theta| \bfX) = \frac{p_{0}(\theta) \prod_{n=1}^{N} p(\bfx_{n} | \theta)}
				      {p(\bfX)}
    \end{equation}
    
    This posterior is in general intractable due to the normalizing constant. 
The idea behind variational inference is to reduce the inference task to an 
optimization problem rather than an integration problem. To do so, it introduces 
a probability distribution $q(\theta)$ from a tractable family $\mcalQ$, 
optimized to approximate the true posterior to an acceptable standard. The 
approximation is found by minimizing a divergence $D[q(\theta)||p(\theta|\bfX)]$ 
between the approximation and the true posterior. For the vast majority of 
divergences, this objective remains intractable as it usually involves computing 
$p(\bfX)$. VI circumvents the issue by considering the equivalent maximization a 
lower-bound (``ELBO,'' short for \emph{evidence lower-bound}) of that objective,
    \begin{equation}
      \label{eq:VI_ELBO}
      \begin{aligned}
	\mcalL_{D}(q, \bfX, &\bfvphi)	\equiv \log p(\bfX|\bfvphi) - D(q(\theta)||p(\theta|\bfX, \bfvphi))	
      \end{aligned}
    \end{equation}
    
    where $D(.||.)$ is a divergence measure and $\mcalL_{D}(.)$
    denotes the objective function associated with $D$.
    

  \subsection{Notable Divergences and their Families}
    \label{subsec:Background/Known}
    A key component for successful variational inference lies in the choice of 
the divergence metric used in Equation~(\ref{eq:VI_ELBO}). A different 
divergence means a different optimization objective and results in the 
approximation having different properties. Over the years, several have been 
proposed.  The review below here does not intend to be exhaustive, but focuses 
only on the divergences of interest in the context of this paper.

    Arguably, the most famous divergence within the VI community is the 
Kullback-Leibler divergence~\cite{jordan1999introduction},
    \begin{equation}
      \label{eq:KL_div}
      D_{KL}(q||p) = \int q(\theta) \log \left(\frac{q(\theta)}{p(\theta)}\right) d\theta.
    \end{equation}
    It offers a relatively simple to optimize objective. However, because the 
KL-divergence considers the log-likelihood ratio $p/q$, it tends to penalize 
more the region where $q>p$ ---i.e, for any given region over-estimating the 
true posterior is penalized more than underestimating it. The approximation 
derived tends to poorly cover regions of small probability in the target 
model~\cite{turner2011two} while focusing on a number of modes according to what 
is allowed by the constraints of $\mathcal Q$. 
    
    To mitigate this issue, efforts have been made to use broader families of 
divergences, where one meta-parameter can be tuned to modify the mass-covering 
behavior of the approximation. In the context of variational inference, the 
alpha-divergence~\cite{amari2012differential}
has been used to develop power EP~\cite{minka2005divergence} and the black-box 
alpha divergence~\cite{hernandez2016black}. In this paper, however, we focus on 
the R{\'e}nyi divergence ~\cite{renyi1961measures,van2014renyi},
    \begin{equation}
      \label{eq:R_div}
      D_{R}^{\alpha}(p||q) = \frac{1}{\alpha-1} \log \int p(\theta)^{\alpha}q(\theta)^{1-\alpha} d\theta,
    \end{equation}
    used in R{\'e}nyi VI~\cite{li2016renyi}. For this family, the meta-parameter 
$\alpha$ can be used to control the influence granted to likelihood ratio $p/q$ 
on the objective in regions of over/under estimation. This flexibility has 
allowed for improvements on traditional VI on complex models, by fine-tuning the 
meta-parameter to the problem~\cite{depeweg2016learning}.
    
    KL-divergence also suffers from the presence of outliers in the training 
data~\cite{ghosh2017generalized}. To perform robust distribution approximation, 
families of divergences such as the beta-divergence~\cite{basu1998robust} have 
been developed  and used to define a pseudo variational 
objective~\cite{ghosh2016robust}. Instead of solving the optimization problem 
defined in Equation~(\ref{eq:VI_ELBO}), they use a surrogate objective 
function motivated by the beta-divergence. In this paper, however, we focus on
    the gamma-divergence~\cite{fujisawa2008robust}, 
    \begin{equation}
      \label{eq:G_div}
      \begin{aligned}
				&D_{\gamma}^{\beta}(p||q) = \frac{1}{\beta(\beta+1)} \log \int 
					p(\theta)^{\beta+1} d\theta \\
			    & + \frac{1}{\beta+1} \log \int q(\theta)^{\beta+1} d\theta
			    - \frac{1}{\beta} \log \int p(\theta)q(\theta)^{\beta} d\theta.
      \end{aligned}
    \end{equation}
    
    In this family, the parameter $\beta$ controls how much importance is 
granted to elements of small probability. The upshot is that in the case the 
data is contaminated with \emph{outliers} -- here interpreted as data points 
contaminated with noise, which are assumed to be spurious and must not be 
covered by the model, although not easy to clean manually in multivariate 
distributions -- then the tail behavior of the model will be 
compromised\footnote{The point being that we should not focus on changing the 
model to accommodate noise, which might not exist out-of-sample, but to change 
the estimator. The difference between estimator and model is common in 
frequentist statistics, with the Bayesian counterpart being less clear at the 
level of generating a posterior distribution. One could consider a measurement 
error model that accounts for noise at training time, to be removed at test 
time, for instance, at the cost of complicating inference. The estimator is 
considered, in our context, as the choices made in the approximation to the 
posterior.}. If the divergence measure is not flexible enough, accommodating 
outliers may have unintended effects elsewhere in the model. 
~\cite{futami2017variational} propose a framework to use the gamma-divergences 
for pseudo VI. Here again their method only proposes a pseudo-Bayesian 
variational updates where the objective does not satisfy 
Equation~(\ref{eq:VI_ELBO}). Despite that they obtain a posterior robust to 
outliers.
    
    As flexible as the divergences defined in Equations~(\ref{eq:R_div}) 
and~(\ref{eq:G_div}) are, they control only either the mass-covering property or 
the robustness property, respectively. The 
AB-divergence~\cite{cichocki2011generalized} allows for both properties to be 
tuned independently, but to the best of our knowledge it has not yet been used 
in the context of variational inference.
    
%

\section{Scale invariant AB Divergence}
  \label{sec:AVdiv}
  In this section, we extend the definition of the scale invariant 
AB-divergence~\cite{cichocki2011generalized} (sAB), as well as defining it for 
continuous distributions. We also describe how it compares to other commonly 
used divergence measures.
  
  \subsection{A two degrees of freedom family of divergences}
    \label{subsec:AVdiv/div}
    
    Under its simplest form, the AB-divergence cannot be used for variational 
inference as it does not provide any computationally tractable form for the loss 
function $\mcalL_{AB}(.)$ as defined in Equation~(\ref{eq:VI_ELBO}) as one 
cannot isolate the terms involving computing the marginal likelihood $p(\bfX)$. 
Detailed computations are available in Appendix~\ref{app_sec:ABVI}. One could 
use the AB-divergence to perform pseudo variational updates as described 
in~\cite{futami2017variational}. However, in that case we would lose the 
guarantees of divergence minimization.
    
    Consider instead, as our primary divergence of interest, the scale invariant 
version of the AB-divergence. This concept was briefly introduced by 
\cite{cichocki2011generalized}:
    \begin{equation}
	\begin{aligned}
	\fontsize{10}{2}
	\label{eq:sABdiv_0}
	&D_{sAB}^{\alpha,\beta}(p||q)\equiv \frac{1}{\beta(\alpha+\beta)} \log \int p(\theta)^{\alpha+\beta} d\theta \\
	  & \quad + \frac{1}{\alpha(\alpha+\beta)} \log \int q(\theta)^{\alpha+\beta} d\theta \\
	  & \quad - \frac{1}{\alpha\beta} \log \int p(\theta)^{\alpha}q(\theta)^{\beta} d\theta,
      \end{aligned}    
    \end{equation}
    for $(\alpha, \beta) \in \dsR^{2}$ such that $\alpha \neq 0$, $\beta \neq 0$ and $\alpha+\beta \neq 0$. 
    
  \subsection{Extension by continuity}
    \label{subsec:AVdiv/extension}    
    In Equation~(\ref{eq:sABdiv_0}), the sAB divergence is not defined
    on the complete $\dsR^{2}$ space. We extend this definition to
    cover all values $(\alpha,\beta) \in \dsR^{2}$ for the purpose of
    comparison with other known divergences, as shown in Table
    \ref{tab:definition}.  Detailed computations are available in
    Appendix~\ref{app_sec:sAB_continuity}.

 \begin{table}[t]
    \begin{equation}
      \label{eq:sAB_div}
      \begin{aligned}
	&D_{sAB}^{\alpha,\beta}(p||q)\equiv\\ 
	&\!\!\begin{cases}
	  \frac{1}{\alpha\beta}  \log \frac{\left( \int p(\theta)^{\alpha+\beta} d\theta\right)^{\frac{\alpha}{\alpha+\beta}}
					      .\left( \int q(\theta)^{\alpha+\beta} d\theta\right)^{\frac{\beta}{\alpha+\beta}}}
					      {\int p(\theta)^{\alpha}q(\theta)^{\beta} d\theta}, \\ \
		\quad\quad\quad\quad\quad\quad\quad\quad\quad\quad \text{for $\alpha \neq 0,\beta \neq 0,\alpha+\beta \neq 0$}  \\  
		  
	  \frac{1}{\alpha^{2}} \left( \log \int \left(\frac{p(\theta)}{q(\theta)}\right)^{\alpha} d\theta 
				  - \int \log \left(\frac{p(\theta)}{q(\theta)}\right)^{\alpha} d\theta \right), \\ 
		\quad\quad\quad\quad\quad\quad\quad\quad\quad\quad\quad \text{for $\alpha = -\beta \neq 0$} \\

	  \frac{1}{\alpha^{2}} \left( \log \frac{\int q(\theta)^{\alpha} d\theta}{\int p(\theta)^{\alpha} d\theta}   
				- \alpha \log \int q(\theta)^{\alpha} \log \frac{q(\theta)}{p(\theta)} d\theta \right),\\
		\quad\quad\quad\quad\quad\quad\quad\quad\quad\quad\quad  \text{for $\alpha \neq 0,\beta = 0$} \\
		
	  \frac{1}{\beta^{2}} \left( \log \frac{\int p(\theta)^{\beta} d\theta}{\int q(\theta)^{\beta} d\theta}   
				- \beta \log \int p(\theta)^{\beta} \log \frac{p(\theta)}{q(\theta)} d\theta \right),\\
		\quad\quad\quad\quad\quad\quad\quad\quad\quad\quad\quad  \text{for $\alpha = 0,\beta \neq 0$} \\
	  
	  \frac{1}{2} \int (\log p(\theta) - \log q(\theta))^{2} d\theta, \quad\quad \text{for $\alpha = 0,\beta = 0$}
	\end{cases}
      \end{aligned}    
    \end{equation}
 \caption{The extended scaled alpha-beta divergence, defined as a function of its
   parameters $\alpha$ and $\beta$ for the entire $\dsR^{2}$.}
 \label{tab:definition}
 \end{table}
 
    For $\alpha=0$ or $\beta=0$, the sAB-divergence reduces to a KL-divergence 
scaled by a power term. For $\alpha=0$ and $\beta=0$, we get a log-transformed 
Euclidean distance~\cite{huang2015log}.
    
    As we will see in Section~\ref{sec:sABVI}, the sAB-divergence can be used in 
the variational inference context.
    
  \subsection{Special cases}
    \label{subsec:AVdiv/special_cases}
    In this section, we describe how some specific choice of parameters 
$(\alpha, \beta)$ simplifies the sAB-divergence into some known divergences or 
families of divergences.
    
    When $\alpha = 0$ and $\beta = 1$ the sAB-divergence reduces down to the 
Kullback-Leibler divergence as defined in Equation~(\ref{eq:KL_div}). By 
symmetry, the reverse KL is obtained for $\alpha = 1$ and $\beta = 0$.
    
    More generally, when $\alpha+\beta=1$, Equation~(\ref{eq:sAB_div}) becomes
    \begin{equation}
      \begin{aligned}
	\label{eq:sAB_div_AB=1}
	&D_{sAB}^{\alpha+\beta=1}(p||q) = \frac{1}{\alpha(\alpha-1)} \log {\int 
p(\theta)^{\alpha}q(\theta)^{1-\alpha} d\theta}, \\
      \end{aligned}    
    \end{equation}
    and the sAB-divergence is proportional to the R{\'e}nyi-divergence defined 
in Equation~(\ref{eq:R_div}).
    
    When $\alpha = 1$ and $\beta \in \dsR$, Equation~(\ref{eq:sAB_div}) becomes
    \begin{equation}
      \begin{aligned}
	\label{eq:sAB_div_A=1_B}
	&D_{sAB}^{\alpha=1,\beta}(p||q) = \frac{1}{\beta(\beta+1)} \log  \int 
p(\theta)^{\beta+1} d\theta \\
					      & + \frac{1}{\beta+1} \log \int q(\theta)^{\beta+1} d\theta
					      - \frac{1}{\beta} \log \int p(\theta)q(\theta)^{\beta} d\theta.
      \end{aligned}    
    \end{equation}
    and the sAB-divergence is equivalent to gamma-divergence.
    
\begin{figure}[h]
      \includegraphics[width=\linewidth]{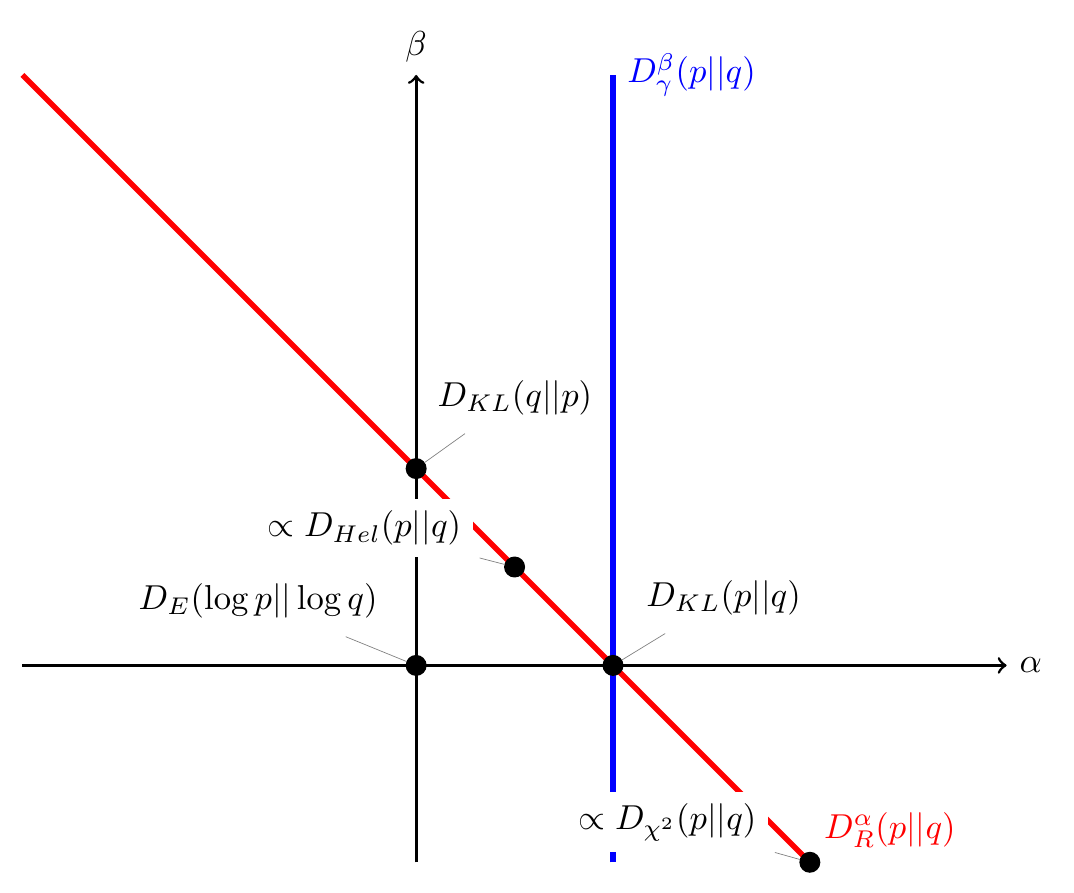}
      \caption{Mapping of the $(\alpha,\beta)$ space. The sAB-divergence reduces 
down to many known divergences but also interpolates smoothly in between them 
and cover a much broader spectrum than the R{\'e}nyi or the gamma-divergence. 
For $(\alpha, \beta)$ equals $(0.5,0.5)$ and $(2,-1)$ the sAB divergence is 
proportional to respectively the Hellinger and the Chi-square divergences. 
Detailed expressions for the divergences mentioned in that Figure are available 
in Appendix~\ref{app_sec:special_cases}.}
      \label{fig:sAB_map}
    \end{figure}

    A mapping of the  $(\alpha, \beta)$ space is shown in 
Figure~\ref{fig:sAB_map}. To summarize, the sAB-divergence allows smooth 
interpolation between many known divergences.
      
  \subsection{Robustness of the divergence}
    \label{subsec:AVdiv/Robustness}
    To develop a better understanding on why using the sAB-divergence might be 
good as a variational objective, we describe how the governing parameters affect 
the optimization problem for various divergences. Let us assume here that the 
approximation $q$ is a function of a vector of parameters $\bfvphi$. Detailed 
computations are available in Appendix~\ref{app_sec:Robustness}.

    Let us first consider as a baseline the usual KL-divergence $D_{KL}(q||p)$. 
Its derivative with regard to $\bfvphi$ is
    \begin{equation}
      \begin{aligned}
      \label{eq:KL_div_grad_0}
      \frac{d}{d\bfvphi} D_{KL}(q||p) = - \int \frac{dq(\theta)}{d\bfvphi} \left( \log \frac{p(\theta)}{q(\theta)} - 1\right) d\theta.
      \end{aligned}    
    \end{equation}
    
    The log-term in Equation~(\ref{eq:KL_div_grad_0}) increases with the cost 
over-estimating $p$ and hence causes the underestimation of the posterior 
variance~\cite{turner2011two}.
    
    In order to gain more flexibility in the approximation behavior, some have 
suggested using broader families of divergences to formulate the variational 
objective. The R{\'e}nyi divergence~\cite{li2016renyi} is one of them and 
differentiating it with regard to $\bfvphi$ yields,
    \begin{equation}
      \begin{aligned}
      \label{eq:R_div_grad_0}
      \frac{d}{d\bfvphi} D_{R}^{\alpha}(q||p) 
	= - \frac{\alpha}{1-\alpha}  \frac{\int \frac{dq(\theta)}{d\bfvphi}\left( \frac{p(\theta)}{q(\theta)}\right) ^{1-\alpha} d\theta}
			                  {\int q(\theta)^{\alpha}p(\theta)^{1-\alpha} d\theta}.
      \end{aligned}    
    \end{equation}
    
    When using the R{\'e}nyi-divergence as an objective, the influence of the 
ratio of $p/q$ is deformed by a factor $\alpha$. This allows the practitioner to 
select whether to emphasize the relative importance of the large ratios (\ie set 
$\alpha < 0$) or on the small ones (\ie set $\alpha > 0$), thus going from
respectively mass-covering to mode-seeking behavior. This does not, however, 
provide any mechanism to handle outliers or rare events.
  
    In the case of the gamma-divergence discussed 
by~\cite{futami2017variational}, its derivative with regard to
    $\bfvphi$ is
    \begin{equation}
      \label{eq:G_div_grad_0}
      \begin{aligned}
	\frac{d}{d\bfvphi} D_{\gamma}^{\beta}(q||p) 
	  = -\frac{1}{\beta}  &\left(  \frac{\int \frac{dq(\theta)}{d\bfvphi} q(\theta)^{\beta} \frac{p(\theta)}{q(\theta)} d\theta}
				            {\int q(\theta)^{\beta}p(\theta) d\theta} \right.\\
	    & \left. - \beta  \frac{\int \frac{dq(\theta)}{d\bfvphi} q(\theta)^{\beta} d\theta}{\int q(\theta)^{\beta+1} d\theta}
		   \right).
      \end{aligned}    
    \end{equation}
  
    When using the gamma-divergence, the influence of the ratio $p/q$ in the 
gradient is weighted by the factor $q(\theta)^{\beta}$. For $\beta<1$, its 
influence is reduced for small values of $q$ causing robustness to outliers. For 
$\beta>1$, the influence of ratios where $q$ is large is reduced instead causing 
a focus on outliers. By setting $\beta$ to  values slightly below $1$, one can 
achieve robustness to outliers whilst maintaining the efficiency of the 
objective~\cite{fujisawa2008robust}.
    
    Finally differentiating the sAB-divergence with regard to $\bfvphi$ yields
    \begin{equation}
      \begin{aligned}
	\label{eq:sAB_div_grad_0}
	&\frac{d}{d\bfvphi} D_{sAB}^{\alpha,\beta}(q||p) = \\
	  & -\frac{1}{\beta} \left(\frac{\int \frac{dq(\theta)}{d\bfvphi} q(\theta)^{\alpha+\beta-1} 
								  \left(\frac{p(\theta)}{q(\theta)}\right)^{\beta} d\theta}
						      {\int q(\theta)^{\alpha}p(\theta)^{\beta} d\theta} \right. \\
	  & \left. \quad\quad\quad - \alpha\beta \frac{\int \frac{dq(\theta)}{d\bfvphi} q(\theta)^{\alpha+\beta-1} d\theta}
					  {\int q(\theta)^{\alpha+\beta} d\theta} \right).
      \end{aligned}    
    \end{equation}
    
\begin{figure}[h]
      \includegraphics[width=\linewidth]{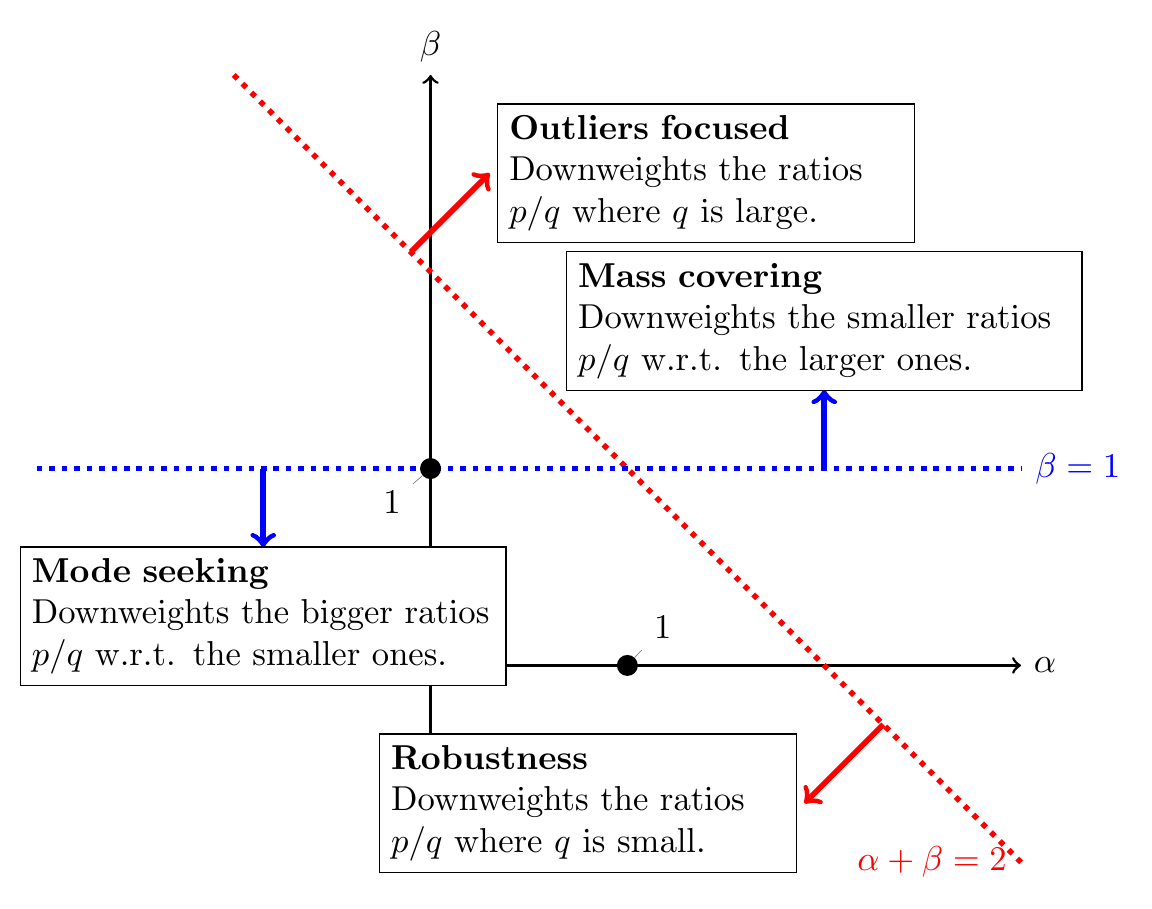}
      \caption{Graphical illustration of the influence of the set control 
parameters $(\alpha, \beta)$. The red line $<\alpha + \beta = 2>$ shows the 
region where the robustness factor $q(\theta)^{\alpha+\beta-1}$ in 
Equation~(\ref{eq:sAB_div_grad_0}) is uniform. The blue line $<\beta = 1 >$ 
shows the region where the ratio $p/q$ in the mass-seeking term 
$(p(\theta)/q(\theta))^{\beta}$ is constant and equal to that of the standard 
Kullback-Leibler divergence.}
      \label{fig:robustness_viz}
    \end{figure}

    The two meta-parameters of the sAB-divergence allow us to combine the 
effects of both the gamma and the R{\'e}nyi divergences. All the terms similar 
to Equation~(\ref{eq:G_div_grad_0}) are controlled by the parameter 
$\alpha+\beta-1$. For the sake of clarity, in the reminder of the paper we will 
use the expression $\lambda=\alpha+\beta$ and parameterize the AB divergence in 
terms of $\lambda$ and $\beta$. One can control the robustness of the objective 
by varying $\lambda$. By setting it to small values below $2$, one can achieve 
robustness to outliers while maintaining the efficiency of the objective.  The 
terms responsible for the ``mode-seeking'' behavior as seen in 
Equation~(\ref{eq:R_div_grad_0}) are here governed by the term $1-\beta$. Thus, 
for $\beta > 1$, one gets the objective to promote a mass-covering behavior. For 
$\beta < 1$, it promotes mode-seeking behaviors.
    Figure~\ref{fig:robustness_viz} provides a visual explanation of the  
influence of each parameters.
        
    In the remainder of the paper, we will report the values used to instantiate 
the sAB-divergence using $\lambda = \alpha+\beta$ instead of $\alpha$ to get a 
direct understanding in terms of robustness and mass covering properties.

\begin{figure}[h]
      \minipage{0.24\textwidth}
	\includegraphics[width=\linewidth]{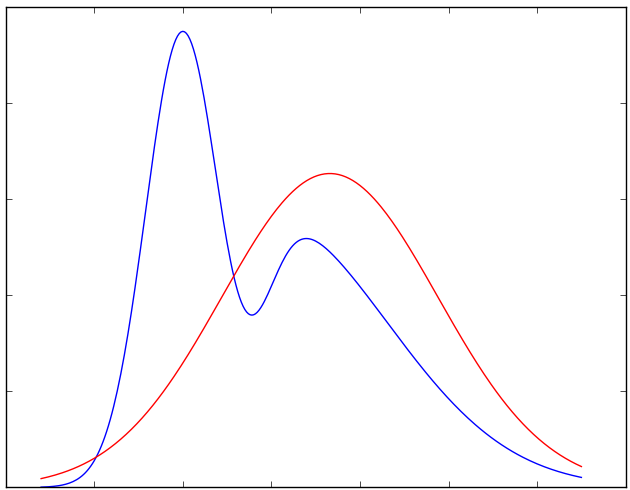}
	\caption*{$\lambda=2.4$, $\beta=-1.0$}      	
	\label{fig:l1.6_b-1.0}
      \endminipage\hfill
      \minipage{0.24\textwidth}
	\includegraphics[width=\linewidth]{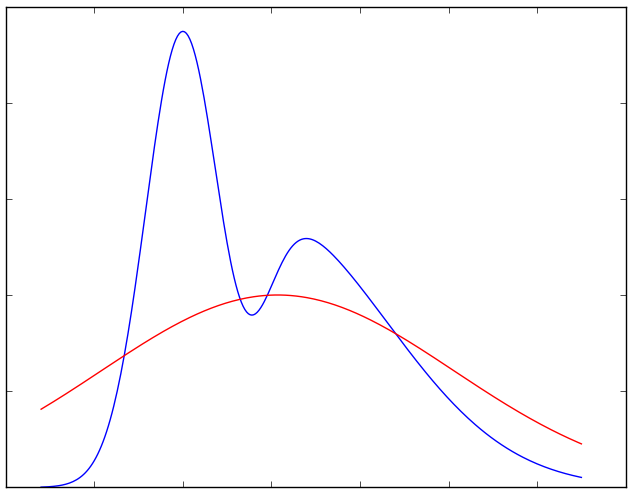}
	\caption*{$\lambda=2.4$, $\beta=2.0$}		
	\label{fig:l1.6_b3.0}
      \endminipage\hfill
      \minipage{0.24\textwidth}%
	\includegraphics[width=\linewidth]{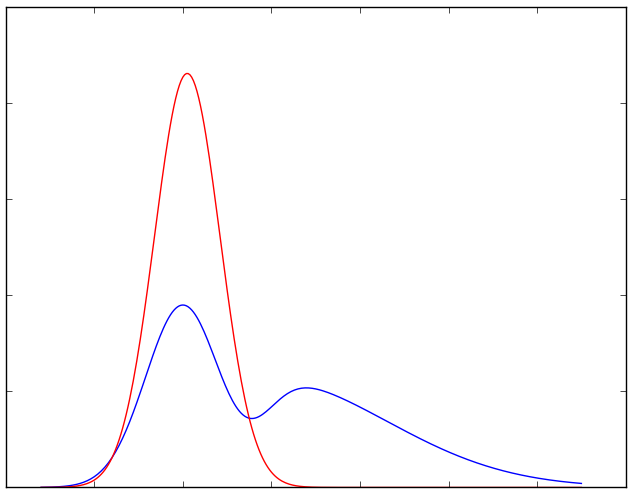}
	\caption*{$\lambda=1.8$, $\beta=-1.0$}		
	\label{fig:l0.4_b-1.0}
      \endminipage\hfill
	\minipage{0.24\textwidth}%
	\includegraphics[width=\linewidth]{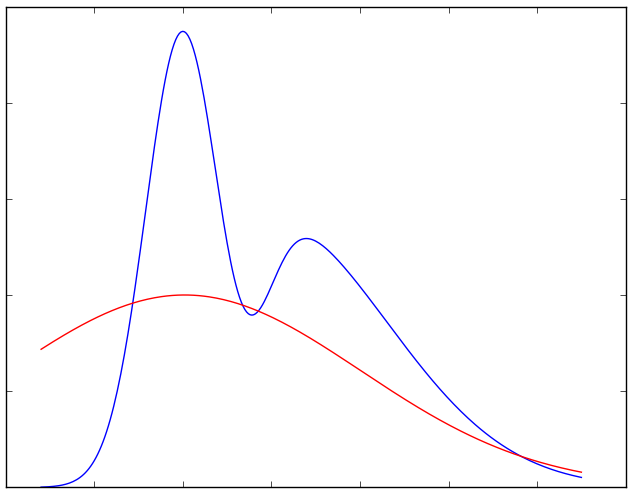}
	\caption*{$\lambda=1.8$, $\beta=2.0$}			
	\label{fig:l0.4_b3.0}
      \endminipage
      \caption{Approximation of a mixture of 2 skewed densities $p$ by a 
Gaussian $q$ for various parameters $\lambda$ and $\beta$.  $\lambda < 2$ causes 
the objective to be robust to outliers, while $\lambda > 2$ increases their 
weight. $\beta>1$ causes the objective to have a mass-covering property, whilst 
$\beta<1$ enforce mode-seeking.}
      \label{fig:sAB_div_illus}
    \end{figure}

    To further illustrate the flexibility offered by the two control parameters 
of the sAB-divergence, Figure~\ref{fig:sAB_div_illus} shows the approximation 
$q$ minimizing $D_{sAB}(\alpha,\beta)(q||p)$. Here $p$ is set to be a mixture of 
two skewed unimodal densities --- a tall and narrow one combined with a short 
and wide density. Density $q$ is required to be a single (non skewed) Gaussian 
with arbitrary mean and variance.
      
    The sAB divergence allows to smoothly tune the properties of the objective 
between ``mass covering'' and ``robustness to outliers.'' In this sense, it is a 
richer objective than either the R{\'e}nyi or the gamma divergences, which can 
only affect respectively the ``mass covering'' or the ``robustness'' properties.

\section{sAB-divergence Variational Inference}
  \label{sec:sABVI}
		Let us consider a posterior distribution of interest $p(\theta| \bfX)$ as 
well as a probability distribution $q(\theta)$ set to approximate the true 
posterior and let us derive the associated sAB variational objective.
  
  \subsection{sAB Variational Objective}
    \label{subsec:sABVI/Objective}
    As seen in Section~\ref{subsec:Background/VI}, the variational approximation 
is fitted by minimizing the divergence between the true distribution and the 
approximated posterior. Using the sAB-divergence defined in 
Equation~(\ref{eq:sAB_div}) we get the following objective,
      \begin{equation}
				\begin{aligned}
					\label{eq:sAB_div_optim_0}
					&D_{sAB}^{\alpha,\beta}(q(\theta)||p(\theta|\bfX)) \\
					  &= \frac{1}{\alpha(\alpha+\beta)} \log \dsE_{q} \left[ 
								\frac{p(\theta, 		\bfX)^{\alpha+\beta}}{q(\theta)} \right] \\
					  &\quad + \frac{1}{\beta(\alpha+\beta)} \log \dsE_{q} \left[ 
			q(\theta)^{\alpha+\beta-1} \right] \\
					  &\quad - \frac{1}{\alpha\beta} \log \dsE_{q} \left[ 
			q(\theta)^{\alpha+\beta-1} \left(\frac{p(\theta, 
			\bfX)}{q(\theta)}\right)^{\beta} \right]
				\end{aligned}    
      \end{equation}
     
    Details of the computation as well as the extension to the complete domain 
of definition are detailed in Appendix~\ref{app_sec:sAB_VI}.
    
    The scale invariant AB-divergence between the true posterior and the 
variational approximation can be expressed as a sum of expectations with regard 
to the variational approximation. Usually in variational inference, the term 
corresponding the marginal likelihood $p(\bfX)$ is dropped, so that the 
objective function is not the divergence itself but an expression that can be 
interpreted as a lower bound on the marginal likelihood, the ELBO. Here, we 
optimize directly on the divergence itself as the terms involving the 
probability of the data $p(\bfX)$ cancel each other. At least in principle, this 
provide a way of directly comparing different choices of $q$ regarding the 
quality of their approximation. This however does not mean that the computation 
of (\ref{eq:sAB_div_optim_0}) can be done exactly, as we will resort to Monte 
Carlo approximations in the next section.
    
    Equation~(\ref{eq:sAB_div_optim_0}) has three main components,
    
    \begin{itemize}
	    \item The first term ensure the objective satisfies the properties of a 
divergence. $D_{sAB}$ is always positive and it is equal to $0$ if and only if 
$p=q$.
      \vspace{-5pt}
      \item The second element and the weighting of the ratio $p(\theta, \bfX) / 
q(\theta)$ in the third element by $q(\theta)^{\alpha+\beta-1}$ control the 
sensibility to outliers As seen in Section~\ref{subsec:AVdiv/Robustness}, by 
setting $\lambda=\alpha+\beta$ to small values below $2$, one can achieve 
robustness to outliers whilst maintaining the efficiency of the objective.
      \vspace{-5pt}
      \item The scaling on the ratio $p(\theta, \bfX) / q(\theta)$ by a power 
$\beta$ in the last element is similar to the bound objective 
of~\cite{li2016renyi} and favors the mass-covering property.
    \end{itemize}
    
    
  \subsection{Optimization framework}
    \label{subsec:sABVI/Optimisation}
    Unfortunately, in general the objective defined in 
Equation~(\ref{eq:sAB_div_optim_0}) still remains intractable and further 
approximations need to be made. As observed in 
Section~\ref{subsec:AVdiv/special_cases}, the sAB-divergence has a form very 
similar to the R{\'e}nyi divergence, so we here use the same approximations as 
in~\cite{li2016renyi}. Theoretically, however, this objective could be used 
with any optimization method as long we are able to compute $p(\theta, \bfX)$ 
and $q(\theta)$ independently (\ie not computing the ratio of the two).
    
    To simplify the computation of the objective, a simple Monte Carlo (MC) 
method is deployed, which uses finite samples $\theta_{k} \sim q(\theta)$, $k = 
1, \dots, K$ to approximate $D_{sAB}^{\alpha,\beta} \approx 
\hat{D}_{sAB}^{\alpha,\beta, K}$,
    
    \begin{equation}
      \begin{aligned}
      \label{eq:sAB_div_MCapprox}
	&\hat{D}_{sAB}^{\alpha,\beta,K}(q(.)||p(.|\bfX) ) \\
        &= \frac{1}{\alpha(\alpha+\beta)} \log \frac{1}{K} \sum_{k=1}^{K}  
							\frac{p(\theta_{k}, \bfX)^{\alpha+\beta}}{q(\theta_{k}| \bfX)} \\
      &\quad + \frac{1}{\beta(\alpha+\beta)} \log \frac{1}{K}  \sum_{k=1}^{K}  
							q(\theta_{k}| \bfX)^{\alpha+\beta-1} \\
      &\quad - \frac{1}{\alpha\beta} \log \frac{1}{K} \sum_{k=1}^{K}  \left[ 
					q(\theta_{k}| \bfX)^{\alpha+\beta-1} \left(\frac{p(\theta_{k}, 
					\bfx)}{q(\theta_{k}| \bfX)}\right)^{\beta} \right].
    \end{aligned}    
  \end{equation}
   
    We also use the reparametrization trick~\cite{kingma2013auto}, along with 
gradient based methods as explained in the next section.
    
\section{Experiments} 
  \label{sec:Exp}
  To demonstrate the advantages of the sAB-divergence over a simpler objective, 
we use it to train variational models on regression tasks on both synthetic and 
real dataset corrupted with outliers. The following experiments have been 
implemented using \textit{tensorflow} 
and \textit{Edward}~\cite{tran2016edward} and the code is publicly available at 
\url{github.com}\footnote{To be release upon publication.}.
  
  \subsection{Regression on synthetic dataset}
    \label{subsec:Exp/Reg}
    First, similarly to~\cite{futami2017variational}, we fit a Bayesian linear 
regression model~\cite{murphy2012machine} to a two-dimensional toy dataset 
where $5\%$ of the data points are corrupted and observe how the generalization 
performances are affected for various training objectives on a non corrupted 
test set. We use a fully factorized Gaussian approximation to the true posterior 
$q(\theta)$. A detailed experimental setup is provided in 
Appendix~\ref{app_sec:Exps}. 
    
\begin{figure}[h]
      \minipage{0.24\textwidth}
	\includegraphics[width=\linewidth]{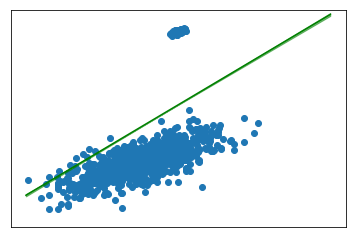}
	\caption*{$\lambda=1.0$ $\beta=0.0$}
	\label{fig:reg_kl}
      \endminipage\hfill
      \minipage{0.24\textwidth}
	\includegraphics[width=\linewidth]{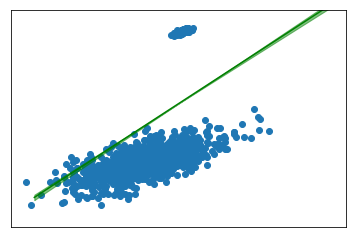}
	\caption*{$\lambda=1.0$ $\beta=0.3$}
	\label{fig:reg_renyi}
      \endminipage\hfill
      \minipage{0.24\textwidth}%
	\includegraphics[width=\linewidth]{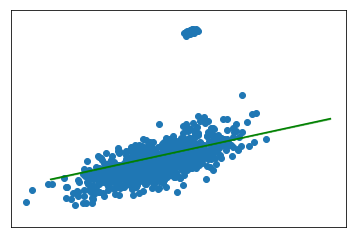}
	\caption*{$\lambda=1.8$ $\beta=0.8$}
	\label{fig:reg_gamma}
      \endminipage\hfill
	\minipage{0.24\textwidth}%
	\includegraphics[width=\linewidth]{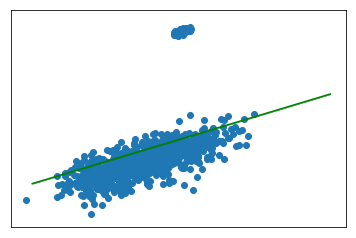}
	\caption*{$\lambda=1.9$ $\beta=-0.3$}
	\label{fig:reg_ab}
      \endminipage
      \caption{ Bayesian linear regression fitted to a dataset containing
        outliers using several sAB objectives. The parameters $\alpha$
        and $\beta$ can be used to ensure robustness to outliers.}
      \label{fig:reg_exp}
    \end{figure}

    The mean of the predictive distributions for various values of 
$(\alpha,\beta)$ are displayed in Figure~\ref{fig:reg_exp} and 
Table~\ref{tab:reg_exp}.  As expected, the network trained with standard VI is 
highly sensitive to outliers and thus has poor predictive abilities at test 
time, where contamination did not happen. On the other end, when trained with 
$(\lambda,\beta)=(1.8, 0.8)$ ---for this values the sAB-divergence is 
equivalent to a gamma distribution set up to be robust to outliers---, the 
predictive distribution ignores the corrupted values. More complex behavior can 
be obtained by tuning the values of the pair $(\alpha,\beta)$ but only yield 
little improvement on such a simple problem.
 
    \begin{table}[h]
      \begin{tabular}{ |p{2.9cm}||p{2cm}|p{2cm}|  }
	\hline
	$(\lambda, \beta)$	& MAE				& MSE		\\
	\hline
	$(1,0, 0.0)$ (KL)  	& $0.58 \pm 0.001$ 		& $0.53 \pm 0.003$ 	\\ 
	$(1.0, 0.3)$ (Renyi)	& $0.58 \pm 0.003$		& $0.51 \pm 0.007$ 	\\
	$(1.8, 0.8)$ (Gamma)	& $0.34 \pm 0.025$		& $0.21 \pm 0.030$ 	\\
	$\mathbf{(1.9,-0.3)}$ \textbf{(sAB)}	& $\mathbf{0.34 \pm 0.025}$	& 
$\mathbf{0.21 \pm 0.030}$ 	\\
	\hline
      \end{tabular}   
      \caption{Average Mean Square Error and Mean Absolute Error over 40 
regression experiments on the same toy dataset where the training data contain a 
$5\%$ proportion of corrupted values.}
      \label{tab:reg_exp}
    \end{table}
    
  \subsection{UCI datasets regression}
    In this section, we show that cross validation can be used to fine-tune the 
parameters $(\alpha, \beta)$ to outperform standard variational inference with a 
KL-objective.
    
    We use here a Bayesian neural network regression 
model~\cite{neal2012bayesian} with Gaussian likelihood on datasets collected 
from the UCI dataset repository~\cite{Lichman2013}. We also artificially corrupt 
part of the outputs in the training data to test the influence of outliers.
    
    For all the experiments, we use a two-layers neural network with $50$ hidden 
units with ReLUs activation functions. We use a fully factorized Gaussian 
approximation to the true posterior $q(\theta)$. Independent standard Gaussian 
priors are given to each of the network weights. The model is optimized using 
ADAM~\cite{kingma2014adam} with learning rate of $0.01$ and the standard 
settings for the other parameters. We perform nested 
cross-validations~\cite{cawley2010over} where the inner validation is used to 
select the optimal parameters $\alpha$ and $\beta$ within the $[-0.5, 
2.5]\times[-1.5,1.5]$ (with step $0.25$). Table~\ref{tab:uci_exp} reports the 
Root Mean Squared Error (RMSE) for the two best pairs $(\alpha, \beta)$ and for 
the KL (i.e. $(\alpha, \beta)=(1,0)$).
    
      \begin{table}[h]
      \begin{tabular}{|p{3.5cm}|p{3.5cm}|  }
	\hline
	$(\alpha+\beta, \beta)$ 		& RMSE				\\
	\hline
	\multicolumn{2}{|c|}{Boston housing - $p_{\text{outliers}}=0\%$}	\\ 
	\hline
	$\mathbf{(1,0, 0.0)}$ \textbf{(KL)} 	& $\mathbf{0.99 \pm 0.031}$	\\ 
	$(1.0, 0.25)$ (sAB)			& $1.01 \pm 0.015$		\\
	$(0.0, -0.75)$ (sAB)			& $1.03 \pm 0.021$		\\
	\hline
	\multicolumn{2}{|c|}{Boston housing - $p_{\text{outliers}}=10\%$}	\\ 
	\hline
	$(1,0, 0.0)$ (KL) 			& $1.13 \pm 0.043$		\\ 
	$\mathbf{(1.25,-0.5)}$ \textbf{(sAB)}	& $\mathbf{1.07 \pm 0.016}$	\\ 
	$(1.75,-0.25)$ (sAB)			& $1.12 \pm 0.029$		\\ 
	\hline
	\multicolumn{2}{|c|}{Concrete - $p_{\text{outliers}}=0\%$}		\\ 
	\hline
	$(1,0, 0.0)$ (KL) 			& $1.01 \pm 0.002$		\\ 
	$\mathbf{(1.0,-1.0)}$ \textbf{(sAB)}	& $\mathbf{0.99 \pm 0.001}$	\\ 
	$(1.5,-0.5)$ (sAB)			& $1.02 \pm 0.003$		\\ 
	\hline
	\multicolumn{2}{|c|}{Concrete - $p_{\text{outliers}}=10\%$}		\\ 
	\hline
	$(1,0, 0.0)$ (KL) 			& $1.16 \pm 0.002$		\\ 
	$\mathbf{(1.5,-0.25)}$ \textbf{(sAB)}	& $\mathbf{1.07 \pm 0.008}$	\\ 
	$(1.25, -0.5)$ (sAB)			& $1.08 \pm 0.003$		\\ 
	\hline
	\multicolumn{2}{|c|}{Yacht - $p_{\text{outliers}}=0\%$}	\\ 
	\hline
	$\mathbf{(1,0, 0.0)}$ \textbf{(KL)} 	& $\mathbf{0.98 \pm 0.021}$	\\ 
	$(1.0, 0.5)$ (sAB)			& $1.00 \pm 0.011$		\\ 
	$(1.0, -1.0)$ (sAB)			& $1.01 \pm 0.003$		\\ 
	\hline
	\multicolumn{2}{|c|}{Yacht - $p_{\text{outliers}}=10\%$}		\\ 
	\hline
	$(1,0, 0.0)$ (KL) 			& $1.09 \pm 0.025$		\\ 
	$\mathbf{(1.25, -0.25})$ \textbf{(sAB)} 	& $\mathbf{1.05 \pm 0.011}$	\\ 
	$(1.75, -0.5)$ (sAB)			& $1.06 \pm 0.017$		\\ 
	\hline
      \end{tabular}   
      \caption{Regression accuracy of a two layer Bayesian neural network 
trained on datasets from the UCI bank of datasets with corrupted by 
$p_{\text{outliers}}\%$ training points. The flexibility offered by the 
sAB-objective allows us to outperform KL-VI in most of the cases where there is 
noise contamination.}
      \label{tab:uci_exp}
    \end{table}   
    
    In the case of uncorrupted data, KL-divergence is often the best choice of 
objective though other set of values for $(\alpha, \beta)$ geared toward mode 
seeking can yield comparable predictive performances. As expected when 
contaminated with outliers, a carefully selected set of parameters such that 
$\alpha+\beta<2$ allows achieving better generalization performances on a non 
corrupted test set compared to VI with KL. In most of the cases ---with and 
without outliers--- the best test score is achieved with $\beta<1$, 
corresponding to a mode-seeking type of objective.    
%
\section{Conclusion}
  \label{sec:Ccl}
  We introduced the extended sAB divergence and its associated variational 
objective. This objective minimize directly the divergence and does not require 
to define an equivalent objective via a lower bound. Furthermore, this family 
of divergence covers most of the already known methods and extend them into a 
more general framework which taps into the growing literature of Monte Carlo 
methods for complex variational objectives. As the resulting objective functions 
are not bounds, they provide a way of directly comparing different 
approximating posterior families, provided that the Monte Carlo error is not 
difficult to control.

	We show that the two governing meta-parameters of the objective allow to 
control independently the mass-covering character and the robustness of the 
approximation. Experimental results point out the interest of this flexible 
objective over the already existing ones for data corrupted with outliers.
  

%
%


\bibliography{2018_icml}
\bibliographystyle{icml2018}

\newpage


\appendix

\onecolumn
\icmltitle{Alpha-Beta Variational Inference - Appendix}

The appendix is organised as follows. Section~\ref{app_sec:ABVI}, we review why 
it was not possible to use the AB-divergence for VI. 
Section~\ref{app_sec:sAB_continuity} develops the computations to extend the 
sAB-divergence by continuity to $(\alpha, \beta) \in \dsR^{2}$. 
Section~\ref{app_sec:Robustness} provides  the mathematical details fo the 
computation of the influence of each parameter. 
Section~\ref{app_sec:special_cases} lists and decribes all the divergences 
encompassed within the sAB-divergence. Section~\ref{app_sec:sAB_VI}, we provide 
a more detailled derivation of the sAB-variational objective. Finally, 
Section~\ref{app_sec:Exps} details the experimental setups used in the core 
paper.\\

\section{AB variational Inference:}
	\label{app_sec:ABVI}
	In the core paper, we use the scale invariant version of the AB-divergence 
(sAB-divergence) to derive the variational objective. We here show why the 
simple AB-divergence cannot be used for this.

	In~\cite{cichocki2011generalized} the AB-divergence is defined as,
  \begin{equation}
    \label{app_eq:AB_div}
    D_{AB}^{\alpha,\beta}(p||q) =
      - \frac{1}{\alpha\beta} \int \left(
      p(\theta)^{\alpha} q(\theta)^{\beta} - 
\frac{\alpha}{\alpha+\beta}p(\theta)^{\alpha+\beta} 
	      - \frac{\beta}{\alpha+\beta}q(\theta)^{\alpha+\beta} \right) d\theta.
  \end{equation}
  
  Let us try to derive the ELBO associated with this divergence,
  \begin{equation*}
    \begin{split}
      D_{AB}^{\alpha, \beta}(q(\theta)||p(\theta|\bfX) )
      &= - \frac{1}{\alpha\beta} \int \left(
							q(\theta)^{\alpha} p(\theta|\bfX) ^{\beta} 
							- \frac{\alpha}{\alpha+\beta}q(\theta)^{\alpha+\beta} 
				      - \frac{\beta}{\alpha+\beta}p(\theta|\bfX) ^{\alpha+\beta} 
						\right) d\theta \\
      &= - \frac{1}{\alpha\beta} \int \left(
					q(\theta)^{\alpha}\left(\frac{p(\theta,\bfX)}{p(\bfX)}\right)^{\beta} 
							- \frac{\alpha}{\alpha+\beta}q(\theta)^{\alpha+\beta} 
				      - 
\frac{\beta}{\alpha+\beta}\left(\frac{p(\theta,\bfX)}{p(\bfX)}\right)^{
\alpha+\beta} 
						\right) d\theta  \\
      &= - \frac{1}{\alpha\beta} \left(
			p(\bfX)^{-\beta} \int q(\theta)^{\alpha}{p(\theta,\bfX)}^{\beta} d\theta 
			- \frac{\alpha}{\alpha+\beta} \int q(\theta)^{\alpha+\beta} d\theta
			- \frac{\beta}{\alpha+\beta} 
			  p(\bfX)^{-(\alpha+\beta)} \int p(\theta,\bfX)^{\alpha+\beta} d\theta
			  \right)
    \end{split}    
  \end{equation*}

	At that step for the KL-divergence or the Renyi-divergence, one can use the
$\log$ term to separate the products in sums and isolate the likelihood of the
data $p(\bfX)$ from the rest of the equation (i.e. the ELBO). For the
AB-divergence, however, we cannot apply this and isolate the
intractable terms. This makes using the AB-divergence for variational inference
impossible. We will see in section~\ref{app_sec:sAB_VI} that this is not the
case for the scale invariant AB-divergence.\\

\section{Extension by continuity of the sAB-divergence}
  \label{app_sec:sAB_continuity}
  We here provide details of the extension by continuity of the sAB-divergence.
	
	In~\cite{cichocki2011generalized} they define the scale invariant 
AB-divergence as,
  \begin{equation}
    \begin{aligned}
    \label{app_eq:sABdiv_1}
    &D_{sAB}^{\alpha,\beta}(p||q)= \frac{1}{\beta(\alpha+\beta)} \log \int p(\theta)^{\alpha+\beta} d\theta \\
	& \quad + \frac{1}{\alpha(\alpha+\beta)} \log \int q(\theta)^{\alpha+\beta} d\theta 
	- \frac{1}{\alpha\beta} \log \int p(\theta)^{\alpha}q(\theta)^{\beta} d\theta,
    \end{aligned}    
  \end{equation}
  for $(\alpha, \beta) \in \dsR^{2}$ such that $\alpha \neq 0$, $\beta \neq 0$ and $\alpha+\beta \neq 0$.\\
  
  We here provide detailed computation of the extension of the domain of definition to $\dsR^{2}$. 
  For simplicity we authorize ourselves to use some shortcuts in the notations 
of undetermined forms.
  
  \subsection{$\alpha+\beta=0$}
    In that case $\beta \rightarrow -\alpha$ and Equation~\ref{app_eq:sABdiv_1} 
		becomes,
    \begin{equation}
      \begin{aligned}
      \label{app_eq:sABdiv_ab->0}
      &D_{sAB}^{\alpha+\beta \rightarrow 0}(p||q) \\
      &= \frac{1}{\beta(\alpha+\beta)} \log \int \left(1 + (\alpha+\beta) \log p(\theta) \right) d\theta 
	  + \frac{1}{\alpha(\alpha+\beta)} \log \int \left(1 + (\alpha+\beta) \log q(\theta)\right) d\theta \\
	  &\quad\quad\quad\quad - \frac{1}{\alpha\beta} \log \int p(\theta)^{\alpha}q(\theta)^{\beta} d\theta \\
      &= \frac{1}{\beta(\alpha+\beta)} \int (\alpha+\beta) \log p(\theta) d\theta 
	  + \frac{1}{\alpha(\alpha+\beta)} \int (\alpha+\beta) \log q(\theta) d\theta \\
	  &\quad\quad\quad\quad - \frac{1}{\alpha\beta} \log \int p(\theta)^{\alpha}q(\theta)^{\beta} d\theta\\
      &= -\frac{1}{\alpha} \int \log p(\theta) d\theta + \frac{1}{\alpha} \int \log q(\theta) d\theta \\
	  &\quad\quad\quad\quad + \frac{1}{\alpha^{2}} \log \int \left( \frac{p(\theta)}{q(\theta)} \right)^{\alpha} d\theta.
      \end{aligned}    
    \end{equation}    
    The first approximation uses $x^{a} = 1 + a \log x$ when $a \approx 0$, the second uses $\log x \approx x$ when $x \rightarrow 1$.\\
    
    So finally we get
    \begin{equation}
      \begin{aligned}
      \label{app_eq:sABdiv_ab=0}
      D_{sAB}^{\alpha+\beta = 0}(p||q)
      = \frac{1}{\alpha^{2}} \left( \log \int \left(\frac{p(\theta)}{q(\theta)}\right)^{\alpha} d\theta 
				  - \int \log \left(\frac{p(\theta)}{q(\theta)}\right)^{\alpha} d\theta \right)
      \end{aligned}    
    \end{equation}   
    
  \subsection{$\alpha=0$ and $\beta \neq 0$}
    In that case Equation~\ref{app_eq:sABdiv_1} becomes,
    \begin{equation}
      \begin{aligned}
      \label{app_eq:sABdiv_a->0}
      &D_{sAB}^{\alpha \rightarrow 0, \beta}(p||q) \\
      &= \frac{1}{\beta^{2}} \log \int p(\theta)^{\beta} d\theta 
	  + \frac{1}{\alpha(\alpha+\beta)} \log \int q(\theta)^{\beta} \left(1 + \alpha \log q(\theta) \right) d\theta \\
	  &\quad\quad\quad\quad - \frac{1}{\alpha\beta} \log \int  q(\theta)^{\beta} \left(1 + \alpha \log p(\theta) \right) d\theta \\
      &= \frac{1}{\beta^{2}} \log \int p(\theta)^{\beta} d\theta 
	  + \frac{1}{\alpha(\alpha+\beta)} \log \int q(\theta)^{\beta} d\theta 
	  + \frac{1}{(\alpha+\beta)} \int q(\theta)^{\beta} \log q(\theta) d\theta \\
	  &\quad\quad\quad\quad - \frac{1}{\alpha\beta} \log \int  q(\theta)^{\beta} d\theta
	  - \frac{1}{\beta} \int  q(\theta)^{\beta} \log p(\theta) d\theta \\
      &= \frac{1}{\beta^{2}} \log \int p(\theta)^{\beta} d\theta 
	  - \frac{1}{\beta(\alpha+\beta)} \log \int q(\theta)^{\beta} d\theta 
	  + \frac{1}{(\alpha+\beta)} \int q(\theta)^{\beta} \log q(\theta) d\theta \\
	  &\quad\quad\quad\quad- \frac{1}{\beta} \int  q(\theta)^{\beta} \log p(\theta) d\theta  
      \end{aligned}    
    \end{equation}    
    The first approximation uses $x^{a} = 1 + a \log x$ when $a \approx 0$, the second uses $\log x \approx x$ when $x \rightarrow 1$.\\
    
    So finally we get
    \begin{equation}
      \begin{aligned}
      \label{app_eq:sABdiv_a=0}
      D_{sAB}^{0, \beta}(p||q) = \frac{1}{\beta^{2}} \left( \log \int p(\theta)^{\beta} d\theta
							    - \log \int q(\theta)^{\beta} d\theta   
						- \beta \log \int q(\theta)^{\beta} \log \frac{p(\theta)}{q(\theta)} d\theta \right)
      \end{aligned}    
    \end{equation}  
  
  \subsection{$\alpha \neq 0$ and $\beta = 0$}
    In that case Equation~\ref{app_eq:sABdiv_1} becomes,
    \begin{equation}
      \begin{aligned}
      \label{app_eq:sABdiv_b->0}
      &D_{sAB}^{\alpha, \beta \rightarrow 0}(p||q) \\
      &= \frac{1}{\beta(\alpha+\beta)} \log \int p(\theta)^{\alpha} \left(1 + \beta \log p(\theta) \right) d\theta 
	    + \frac{1}{\alpha^{2}} \log \int q(\theta)^{\alpha} d\theta \\
	  &\quad\quad\quad\quad - \frac{1}{\alpha\beta} \log \int p(\theta)^{\alpha} \left(1 + \beta \log q(\theta) \right) d\theta \\
      &= \frac{1}{\beta(\alpha+\beta)} \log \int p(\theta)^{\alpha} d\theta 
	  + \frac{1}{(\alpha+\beta)} \int p(\theta)^{\alpha} \log p(\theta) d\theta 
	  + \frac{1}{\alpha^{2}} \log \int q(\theta)^{\alpha} d\theta  \\
	  &\quad\quad\quad\quad - \frac{1}{\alpha\beta} \log \int  p(\theta)^{\alpha} d\theta
	  - \frac{1}{\alpha} \int  p(\theta)^{\alpha} \log q(\theta) d\theta \\
      &= -\frac{1}{\alpha(\alpha+\beta)} \log \int p(\theta)^{\alpha} d\theta 
	  + \frac{1}{(\alpha+\beta)} \int p(\theta)^{\alpha} \log (\theta) d\theta
	  + \frac{1}{\alpha^{2}} \log \int q(\theta)^{\alpha} d\theta \\
	  &\quad\quad\quad\quad - \frac{1}{\alpha} \int  p(\theta)^{\alpha} \log q(\theta) d\theta
      \end{aligned}    
    \end{equation}    
    The first approximation uses $x^{a} = 1 + a \log x$ when $a \approx 0$, the second uses $\log x \approx x$ when $x \rightarrow 1$.\\
    
    So finally we get
    \begin{equation}
      \begin{aligned}
      \label{app_eq:sABdiv_a=0}
      D_{sAB}^{\alpha, 0}(p||q) = \frac{1}{\alpha^{2}} \left( \log \int q(\theta)^{\alpha} d\theta
							    - \log \int p(\theta)^{\alpha} d\theta   
						- \alpha \log \int pq(\theta)^{\alpha} \log \frac{q(\theta)}{p(\theta)} d\theta \right)
      \end{aligned}    
    \end{equation}    
    
  \subsection{$\alpha=0$ and $\beta = 0$}
    In that case Equation~\ref{app_eq:sABdiv_1} becomes,
    \begin{equation}
      \begin{aligned}
      \label{app_eq:sABdiv_a->0b->0}
      &D_{sAB}^{\alpha \rightarrow 0, \beta \rightarrow 0}(p||q) \\
      &= \frac{1}{\beta(\alpha+\beta)} \log \int (1 + (\alpha+\beta)  \log p(\theta)) d\theta 
	  + \frac{1}{\alpha(\alpha+\beta)} \log \int (1 + (\alpha+\beta) \log q(\theta)) d\theta \\
	  &\quad\quad\quad\quad - \frac{1}{\alpha\beta} \log \int (1 + \alpha \log p(\theta))(1+ \beta \log q(\theta)) d\theta \\
      &= \frac{1}{\beta(\alpha+\beta)} \log \int (1 + (\alpha+\beta)  \log p(\theta)) d\theta 
	  + \frac{1}{\alpha(\alpha+\beta)} \log \int (1 + (\alpha+\beta) \log q(\theta)) d\theta \\
	  &\quad\quad\quad\quad - \frac{1}{\alpha\beta} \log \int (1 + \alpha \log p(\theta) + \beta \log q(\theta) 
		    + \alpha \beta \log p(\theta) \log q(\theta)) d\theta \\
      &= \frac{1}{\beta(\alpha+\beta)} \int (\alpha+\beta)  \log p(\theta) d\theta 
	  + \frac{1}{\alpha(\alpha+\beta)} \int (\alpha+\beta) \log q(\theta) d\theta \\
	  &\quad\quad\quad\quad - \frac{1}{\alpha\beta} \int (\alpha \log p(\theta) + \beta \log q(\theta) 
		    + \alpha \beta \log p(\theta) \log q(\theta)) d\theta \\
      &= - \int \log p(\theta) \log q(\theta) d\theta   
      \end{aligned}    
    \end{equation}    
    The first approximation uses  $x^{a} = 1 + a \log x$ when $a \approx 0$, the second uses $\log x \approx x$ when $x \rightarrow 1$.\\
    
    So finally we get
    \begin{equation}
      \begin{aligned}
      \label{app_eq:sABdiv_a=0b=0}
      D_{sAB}^{0, 0}(p||q) = \frac{1}{2} \int (\log p(\theta) - \log q(\theta))^{2} d\theta 
      \end{aligned}    
    \end{equation}    
    
\section{Special cases of the sAB-divergence}
	\label{app_sec:special_cases}
	We here provide a more complete list of the known divergences included in the 
	sAB-divergence.\\
		
	For $(\alpha, \beta) = (1,0)$, the sAB-divergence reduces down to the 
	KL-divergence~\cite{kullback1951information},
  \begin{equation}
    \label{app_eq:KL_div}
    D_{sAB}^{(1,0)}(q||p) = \int q(\theta) \log \left( 
							\frac{q(\theta)}{p(\theta)} \right) d\theta.
  \end{equation}

 	For $(\alpha, \beta) = (0,1)$, the sAB-divergence reduces down to the 
	reverse KL-divergence,
  \begin{equation}
    \label{app_eq:rKL_div}
    D_{sAB}^{(1,0)}(q||p) = \int p(\theta) \log \left( 
							\frac{p(\theta)}{q(\theta)} \right) d\theta.
  \end{equation}
  
  For $(\alpha, \beta) = (0.5,0.5)$, the sAB-divergence is a function of the 
  Hellinger-distance~\cite{lindsay1994efficiency},
  \begin{equation}
    \label{app_eq:Hel_div}
    \begin{aligned}
	    D_{sAB}^{(0.5,0.5)}(q||p) 
		    &= -4 \log \int \sqrt{p(\theta)}.\sqrt{q(\theta)}d\theta \\
		    &= -4 \log \int \left(1 - \frac{1}{2} 
					\left(\sqrt{p(\theta)}-\sqrt{q(\theta)}\right)^{2}\right)d\theta \\
		    &= -4 \log (1 - D_{H}(p||q))
		\end{aligned}
  \end{equation}
  
  For $(\alpha, \beta) = (2,-1)$, the sAB-divergence is a function of the 
  $\chi^{2}$-divergence~\cite{nielsen2014chi},
  \begin{equation}
    \label{app_eq:chi_div}
    \begin{aligned}
	    D_{sAB}^{(2,-1)}(q||p) 
		    &= \frac{1}{2} \log \int \frac{p(\theta)^{2}}{q(\theta)}d\theta \\
		    &= \frac{1}{2} \log (1-D_{\chi^{2}}(p||q))
		\end{aligned}
  \end{equation}
  
  For $(\alpha, \beta) = (0, 0)$, the sAB-divergence is equal to the 
log-euclidean divergence $D_{E}$~\cite{huang2015log},
  \begin{equation}
    \label{app_eq:euclidean_div}
    D_{sAB}^{0, 0}(p||q) = \frac{1}{2} \int (\log p(\theta) - \log 
				q(\theta))^{2} d\theta 
  \end{equation}
  
    When $\alpha+\beta=1$, the sAB-divergence is proportional to the 
		R{\'e}nyi-divergence~\cite{renyi1961measures}
    \begin{equation}
			\label{app_eq:sAB_div_AB=1}
			D_{sAB}^{\alpha+\beta=1}(p||q) = \frac{1}{\alpha(\alpha-1)} \log {\int 
					p(\theta)^{\alpha}q(\theta)^{1-\alpha} d\theta}.
    \end{equation}
  
    When $\alpha = 1$ and $\beta \in \dsR$, the sAB-divergence is equivalent to 
		gamma-divergence~\cite{fujisawa2008robust},
    \begin{equation}
			\label{aap_eq:sAB_div_A=1_B}
			D_{sAB}^{\alpha=1,\beta}(p||q) = \frac{1}{\beta(\beta+1)} \log \int 
								p(\theta)^{\beta+1} d\theta \\
					      + \frac{1}{\beta+1} \log \int q(\theta)^{\beta+1} d\theta
					      - \frac{1}{\beta} \log \int p(\theta)q(\theta)^{\beta} d\theta.
    \end{equation}\\
     
\section{Robustness the sAB-divergence}
  \label{app_sec:Robustness}
  We here provide detailed computation of the derivative of various divergences 
with regard to the governing parameters of the approximation. Let us here assume 
we approximate the distribution $p$ by $q$ a function of the vector of 
parameters $\bfvphi$.
  
  \subsection{Kullback-Leibler divergence}
    For the Kullback-Leibler divergence, we get the following results,
   \begin{equation}
      \begin{aligned}
      \label{app_eq:KL_div_grad}     
	\frac{d}{d\bfvphi} D_{KL}(q||p) 
	&= - \frac{d}{d\bfvphi} \left(  \int q(\theta) \log \frac{p(\theta)}{q(\theta)} d\theta \right) \\
	&= -   \int \left( \frac{dq(\theta)}{d\bfvphi} \log \frac{p(\theta)}{q(\theta)} 
		  + q(\theta) \frac{d}{d\bfvphi} \log \frac{p(\theta)}{q(\theta)} \right) d\theta  \\	
	&= - \int \frac{dq(\theta)}{d\bfvphi} \left( \log \frac{p(\theta)}{q(\theta)} - 1\right) d\theta.	
      \end{aligned}    
    \end{equation}
  
  \subsection{R{\'e}nyi-divergence}
    For the R{\'e}nyi-divergence, we get the following results,
    \begin{equation}
      \begin{aligned}
      \label{app_eq:R_div_grad}
      \frac{d}{d\bfvphi} D_{R}^{\alpha}(q||p) 
	&=  - \frac{d}{d\bfvphi} \left( \frac{1}{\alpha-1} \log \int q(\theta)^{\alpha} p(\theta)^{1-\alpha} d\theta \right)\\
	&=  -  \frac{1}{\alpha-1} \frac{\int \frac{dq(\theta)}{d\bfvphi} \alpha q(\theta)^{\alpha-1} p(\theta)^{1-\alpha}}
				       {\int q(\theta)^{\alpha} p(\theta)^{1-\alpha} d\theta} \\				       
	&= - \frac{\alpha}{1-\alpha}  \frac{\int \frac{dq(\theta)}{d\bfvphi}\left( \frac{p(\theta)}{q(\theta)}\right) ^{1-\alpha} d\theta}
			                  {\int q(\theta)^{\alpha}p(\theta)^{1-\alpha} d\theta}.
      \end{aligned}    
    \end{equation}
  
  \subsection{Gamma-divergence}
  For the Gamma-divergence, we get the following results,  
    \begin{equation}
      \begin{aligned}
	\label{app_eq:G_div_grad}
	\frac{d}{d\bfvphi} D_{\gamma}^{\beta}(q||p) 
	  &= \frac{d}{d\bfvphi} \left( \frac{1}{1+\beta} \log \int q(\theta)^{\beta+1} d\theta \right.
				     \left. + \frac{1}{\beta(1+\beta)} \log \int p(\theta)^{\beta+1} d\theta \right. \\
	  & \quad\quad\quad\quad\quad\quad\quad \left. - \log \int q(\theta)^{\beta}p(\theta) d\theta \right) \\
	  &= \frac{1}{1+\beta} \frac{\frac{d}{d\bfvphi} \int q(\theta)^{\beta+1} d\theta}{\int q(\theta)^{\beta+1} d\theta} 
				     - \frac{\frac{d}{d\bfvphi} \int q(\theta)^{\beta}p(\theta) d\theta}{\int q(\theta)^{\beta}p(\theta) d\theta} \\
	  &= \frac{\int \frac{dq(\theta)}{d\bfvphi}q(\theta)^{\beta} d\theta}{\int q(\theta)^{\beta+1} d\theta} 
				     - \beta \frac{\int \frac{dq(\theta)}{d\bfvphi}q(\theta)^{\beta-1}p(\theta) d\theta}{\int q(\theta)^{\beta}p(\theta) d\theta} \\
	  &= - \frac{1}{\beta}  \left( \frac{\int \frac{dq(\theta)}{d\bfvphi} q(\theta)^{\beta} \frac{p(\theta)}{q(\theta)} d\theta}
				            {\int q(\theta)^{\beta}p(\theta) d\theta} 
				        - \beta \frac{\int \frac{dq(\theta)}{d\bfvphi} q(\theta)^{\beta} d\theta}{\int q(\theta)^{\beta+1} d\theta}
				 \right).				            
      \end{aligned}    
    \end{equation}
  
  \subsection{sAB-divergence}
    For the sAB-divergence, we get the following results,
    \begin{equation}
      \begin{aligned}
	\label{app_eq:sAB_div_grad}
	\frac{d}{d\bfvphi} D_{sAB}^{\alpha,\beta}(q||p) 
	  &= \frac{d}{d\bfvphi} \left( \frac{1}{\beta(\alpha+\beta)} \log \int q(\theta)^{\alpha+\beta} d\theta \right.
				     \left. + \frac{1}{\alpha(\alpha+\beta)} \log \int p(\theta)^{\alpha+\beta} d\theta \right. \\
	  & \quad\quad\quad\quad\quad\quad\quad \left. - \log \int q(\theta)^{\alpha}p(\theta)^{\beta} d\theta \right) \\	  
	  &= \frac{1}{\beta(\alpha+\beta)} \frac{\frac{d}{d\bfvphi} \int q(\theta)^{\alpha+\beta} d\theta}{\int q(\theta)^{\alpha+\beta} d\theta} 
				     - \frac{\frac{d}{d\bfvphi} \int q(\theta)^{\alpha}p(\theta)^{\beta} d\theta}
					    {\int q(\theta)^{\alpha}p(\theta)^{\beta} d\theta} \\
	  &= \frac{1}{\beta} \frac{\int \frac{dq(\theta)}{d\bfvphi}q(\theta)^{\alpha+\beta-1} d\theta}{\int q(\theta)^{\alpha+\beta} d\theta} 
				     - \alpha \frac{\int \frac{dq(\theta)}{d\bfvphi}q(\theta)^{\alpha-1}p(\theta)^{\beta} d\theta}
					    {\int q(\theta)^{\alpha}p(\theta)^{\beta} d\theta} \\
	  &= -\frac{1}{\beta} \left(\frac{\int \frac{dq(\theta)}{d\bfvphi} q(\theta)^{\alpha+\beta-1} 
								  \left(\frac{p(\theta)}{q(\theta)}\right)^{\beta} d\theta}
						      {\int q(\theta)^{\alpha}p(\theta)^{\beta} d\theta}
					  - \alpha\beta \frac{\int \frac{dq(\theta)}{d\bfvphi} q(\theta)^{\alpha+\beta-1} d\theta}
						 {\int q(\theta)^{\alpha+\beta} d\theta} \right)
				  .
      \end{aligned}    
    \end{equation}

\section{sAB-divergence Variational Inference}
  \label{app_sec:sAB_VI}
  We here provide detailed computation of the variational objective using the 
sAB-divergence. We also detail the extension of this objective to the complete 
domain of definition.
  
  \subsection{sAB variational objective}
    We are interested in minimizing the divergence $D_{sAB}^{\alpha,\beta}(q(\theta)||p(\theta|\bfX))$, this yields,
    \begin{equation}
      \begin{aligned}      
	&D_{sAB}^{\alpha,\beta}(q(\theta)||p(\theta|\bfX) ) \\
	    &= \frac{1}{\alpha\beta} \log \frac{\left( \int q(\theta)^{\alpha+\beta} d\theta \right)^{\frac{\alpha}{\alpha+\beta}}
					    .\left( \int p(\theta|\bfX)^{\alpha+\beta} d\theta \right)^{\frac{\beta}{\alpha+\beta}}}
					    {\int q(\theta)^{\alpha}p(\theta|\bfX)^{\beta} d\theta}. \\ 
	    &= \frac{1}{\alpha\beta} \left[
	      \log \left( \int q(\theta)^{\alpha+\beta} d\theta \right)^{\frac{\alpha}{\alpha+\beta}}
	      + \log \left( \int p(\theta|\bfX)^{\alpha+\beta} d\theta \right)^{\frac{\beta}{\alpha+\beta}} \right. \\
	       &\quad\quad\quad\quad \left. - \log \left( \int q(\theta)^{\alpha}p(\theta|\bfX)^{\beta} d\theta \right) \right] \\
	    &= \frac{1}{\alpha\beta} \left[
	      \log \left( \int q(\theta)^{\alpha+\beta} d\theta \right)^{\frac{\alpha}{\alpha+\beta}}
	      + \log \left( \int \left(\frac{p(\theta, \bfX)}{p(\bfX)} \right)^{\alpha+\beta} d\theta \right)^{\frac{\beta}{\alpha+\beta}} \right. \\
	      &\quad\quad\quad\quad \left. - \log \left( \int q(\theta)^{\alpha}\left(\frac{p(\theta, \bfX)}{p(\bfX)} \right)^{\beta} d\theta \right) 
	      \right] \\
	    &= \frac{1}{\alpha\beta} \left[
	      \log \left( \int q(\theta)^{\alpha+\beta} d\theta \right)^{\frac{\alpha}{\alpha+\beta}}
	      + \log \left( p(\bfX)^{-(\alpha+\beta)} \int p(\theta, \bfX)^{\alpha+\beta} d\theta \right)^{\frac{\beta}{\alpha+\beta}} \right. \\
	      &\quad\quad\quad\quad \left. - \log \left(p(\bfX)^{-\beta} \int q(\theta)^{\alpha}p(\theta, \bfX)^{\beta} d\theta \right) 
	      \right] \\
	    &= \frac{1}{\alpha\beta} \left[
	      \log \left( \int q(\theta)^{\alpha+\beta} d\theta \right)^{\frac{\alpha}{\alpha+\beta}} \right.
	      \left. + \log \left( \int p(\theta, \bfX)^{\alpha+\beta} d\theta \right)^{\frac{\beta}{\alpha+\beta}} \right. \\
	      &\quad\quad\quad\quad \left.  - \beta \log p(\bfX)  + \beta \log p(\bfX) \right.
	      \left. - \log \left(\int q(\theta)^{\alpha}p(\theta, \bfX)^{\beta} d\theta \right)
	      \right] \\
	    &= \frac{1}{\beta(\alpha+\beta)} \log \int q(\theta)^{\alpha+\beta} d\theta 
	      + \frac{1}{\alpha(\alpha+\beta)}\log \int p(\theta, \bfX)^{\alpha+\beta} d\theta\\
	      &\quad\quad\quad\quad- \frac{1}{\alpha\beta} \log \int q(\theta)^{\alpha}p(\theta, \bfX)^{\beta} d\theta 
	      \end{aligned}    
      \end{equation}
      
      Finally rewriting this expression to make expectations over $q(\theta)$ appears yields,
      \begin{equation}
	\begin{aligned}
	\label{app_eq:sAB_div_optim}
	&D_{sAB}^{\alpha,\beta}(q(\theta)||p(\theta|\bfX)) \\
	  &= \frac{1}{\alpha(\alpha+\beta)} \log \dsE_{q} \left[ \frac{p(\theta, \bfX)^{\alpha+\beta}}{q(\theta)} \right]
	  + \frac{1}{\beta(\alpha+\beta)} \log \dsE_{q} \left[ q(\theta)^{\alpha+\beta-1} \right] \\
	  &\quad\quad\quad\quad - \frac{1}{\alpha\beta} \log \dsE_{q} \left[ \frac{p(\theta, \bfX)^{\beta}}{q(\theta)^{1-\alpha}} \right]
	\end{aligned}    
      \end{equation}

  \subsection{Extension by continuity}
    Computation very similar to those in Section~\ref{app_sec:sAB_continuity} yields,
    \begin{equation}
      \label{app_eq:sAB_VI}
      \begin{aligned}
	&D_{sAB}^{\alpha,\beta}(q(\theta)||p(\theta|\bfX))=\\ 
	&\begin{cases}
	  \frac{1}{\beta(\alpha+\beta)} \log \int q(\theta)^{\alpha+\beta} d\theta 
	      + \frac{1}{\alpha(\alpha+\beta)}\log \int p(\theta, \bfX)^{\alpha+\beta} d\theta 
	      - \frac{1}{\alpha\beta} \log \int q(\theta)^{\alpha}p(\theta, \bfX)^{\beta} d\theta \\
	      \quad\quad\quad\quad\quad\quad\quad\quad\quad\quad\quad\quad\quad\quad\quad\quad\quad\quad\quad\quad\quad\quad\quad
	      \text{for $\alpha,\beta,\alpha+\beta \neq 0$}  \\  
		  
	  \frac{1}{\alpha^{2}} \left( \log \int \left(\frac{q(\theta)}{p(\theta, \bfX)}\right)^{\alpha} d\theta 
				  - \int \log \left(\frac{q(\theta)}{p(\theta, \bfX)}\right)^{\alpha} d\theta \right) 
				  \quad\quad\quad\quad \text{for $\alpha = -\beta \neq 0$} \\
  
	  \frac{1}{\alpha^{2}} \left( \log \int p(\theta, \bfX)^{\alpha} d\theta
				      - \log \int q(\theta)^{\alpha} d\theta   
				      - \alpha \log \int p(\theta, \bfX)^{\alpha} \log \frac{p(\theta, \bfX)}{q(\theta)} d\theta \right)	\\	
	      \quad\quad\quad\quad\quad\quad\quad\quad\quad\quad\quad\quad\quad\quad\quad\quad\quad\quad\quad\quad\quad\quad\quad \text{for $\alpha \neq 0,\beta = 0$} \\

	  \frac{1}{\beta^{2}} \left( \log \int q(\theta)^{\beta} d\theta
				     - \log \int p(\theta, \bfX)^{\beta} d\theta   
				     - \beta \log \int q(\theta)^{\beta} \log \frac{q(\theta)}{p(\theta, \bfX)} d\theta \right)	\\	
	      \quad\quad\quad\quad\quad\quad\quad\quad\quad\quad\quad\quad\quad\quad\quad\quad\quad\quad\quad\quad\quad\quad\quad \text{for $\alpha = 0,\beta \neq 0$} \\
	  
	  \frac{1}{2} \int (\log q(\theta) - \log p(\theta, \bfX))^{2} d\theta, 
				  \quad\quad\quad\quad\quad\quad\quad\quad\quad\quad\quad \text{for $\alpha,\beta = 0$}
	\end{cases}
      \end{aligned}    
    \end{equation}

\section{Experiments}
  \label{app_sec:Exps}
  We here provide a more detailed description of our experimental setups.
  The following experiments have been implemented using 
  \textit{tensorflow}~\cite{tensorflow2015-whitepaper} and 
  \textit{Edward}~\cite{tran2016edward}.
  
  \subsection{Regression on synthetic dataset}
    In this experiment we create a toy dataset to showcase the strength of the 
    sAB variational objective.\\
    
    The non-corrupted data are generated by the following process,
    \begin{equation}
	    y = \bfw \bfX + \mcalN(0,0.1) 
    \end{equation}
		with $\bfw=[1/2...1/2]$ a $D$-dimensional vector and $bfX$ a set of 
		points randomly distributed between $[-1, 1]^{D}$.\\
		A given percentage $p_{outliers}$ of the data are corrupted and follows the 
		process,
		\begin{equation}
			y = 5 + \bfw \bfX + \mcalN(0,0.1)
		\end{equation}
		with $\bfw=[1/2...1/2]$ and $\bfX$ is sampled from $\mcalN(0, 0.2)$.\\
		
		For N such data points $[(\bfx_{n}, y_{n})]_{n \in [1, N]}$, we uses the 
		following distributions,
		\begin{equation}
			\begin{aligned} 
				p(\bf{w}) &= \mcalN(\bfw \mid \mathbf{0}, \sigma_{w}^{2} \mcalI_{D}), \\
				p(b) &= \mcalN(b \mid 0, \sigma_{b}^{2}),
			\end{aligned}
		\end{equation}
		and
		\begin{equation}
			p(y \mid \bfw, b, \bfX) = \prod_{n=1}^{N} \mcalN(y_{n} \
							\mid \bfx_{n}^{\top} \bfw + b, \sigma_{y}^{2}).
		\end{equation}
		We define the variational model to be a fully factorized normal across 
		the weights.\\
		
		For the experiments presented in the paper we use $N=1000$, $D=4$ and 
		$p_{outliers}=5\%$.\\
			
		We train the model using ADAM~\cite{kingma2014adam} with learning rate of 
    $0.01$ for $1000$ steps. We use $5$ MC samples to evaluate the divergence.
		       
  \subsection{UCI datasets regression}
	  We use here a Bayesian neural network regression model with Gaussian 
likelihood on datasets collected from the UCI dataset 
repository~\cite{Lichman2013}. We also artificially corrupt part of the outputs 
in the training data to test the influence of outliers. The corruption is 
achieved by randomly adding $5$ standard deviation to $p_{outliers}\%$ of the 
points after normalization.
    
    For all the experiments, we use a two-layers neural network with $50$ hidden 
units with ReLUs activation functions. We use a fully factorized Gaussian 
approximation to the true posterior $q(\theta)$. Independent standard Gaussian 
priors are given to each of the network weights. The model is optimized using 
ADAM~\cite{kingma2014adam} with learning rate of $0.01$ and the standard 
settings for the other parameters for $500$ epochs. We perform  nested 
cross-validations~\cite{cawley2010over} where the inner validation is used to 
select the optimal parameters $\alpha$ and $\beta$ within the $[-0.5, 
2.5]\times[-1.5,1.5]$ (with step $0.25$). The best model selected from the inner 
loop is then re-trained on the complete outer split. We use $25$ MC samples to 
evaluate the divergence. The outer cross validation used $K_{1})=10$ 
folds and the inner one uses $K_{2})=2$ folds.

\clearpage
\bibliographystyle{apalike}
\bibliography{supplementary/bib_icml_appendix}
\nocite{*}

%
%
%

\end{document}